\pdfoutput=1

\documentclass[11pt]{article}

\usepackage{emnlp2021}

\usepackage{times}
\usepackage{latexsym}

\usepackage{graphicx}
\usepackage{amsmath}
\usepackage{wrapfig}
\newcommand{\figref}[1]{Figure~\ref{#1}}
\usepackage{bbm}
\usepackage{mathtools}
\usepackage{linguex}
\usepackage{csvsimple}
\usepackage[plainruled,linesnumbered,noend]{algorithm2e}
\usepackage[noend]{algpseudocode}
\usepackage{multicol}
\algrenewcommand\algorithmicindent{0em}
\makeatletter
\algrenewcommand\ALG@beginalgorithmic{\small}
\makeatother

\RestyleAlgo{ruled}

\SetKwComment{Comment}{/* }{ */}
\newlength\mylen
\newcommand\myinput[1]{%
  \settowidth\mylen{\KwIn{}}%
  \setlength\hangindent{\mylen}%
  \hspace*{\mylen}#1\\}
  
\makeatletter
\newcommand{\removelatexerror}{\let\@latex@error\@gobble}
\makeatother

\let\oldnl\nl
\newcommand{\nonl}{\renewcommand{\nl}{\let\nl\oldnl}}

\usepackage{caption}
\DeclareCaptionLabelFormat{noname}{#2}
\usepackage[T1]{fontenc}

\usepackage[utf8]{inputenc}

\usepackage{microtype}

\newcommand\blfootnote[1]{%
  \begingroup
  \renewcommand\thefootnote{}\footnote{#1}%
  \addtocounter{footnote}{-1}%
  \endgroup
}
\SetKwIF{If}{ElseIf}{Else}{if}{}{else if}{else}{end if}%

\SetCommentSty{mycommfont}
\title{Probing for Incremental Parse States in Autoregressive  Language Models}

\author{Tiwalayo Eisape$^{1}$
\And
  Vineet Gangireddy$^{2}$
    \And
  Roger P.\ Levy$^{1}$ \And
  Yoon Kim$^{1}$\AND \vspace{-13mm}\\
  MIT$^{1}$, Harvard University$^{2}$\\
\texttt{\{eisape,rplevy,yoonkim\}@mit.edu}\\ \texttt{vineetgangireddy@college.harvard.edu}
}

\begin{document}

\maketitle
\vspace{-15mm}
\begin{abstract}
\vspace{-1mm}
Next-word predictions from autoregressive neural language models show remarkable sensitivity to syntax. This work evaluates the extent to which this behavior arises as a result of a learned ability to maintain implicit representations of incremental syntactic structures. We extend work in syntactic probing to the incremental setting and present several probes for extracting incomplete syntactic structure (operationalized through  parse states from a stack-based parser) from autoregressive language models. We find that our probes can be used to predict model preferences on ambiguous sentence prefixes and causally intervene on model representations and steer model behavior. This suggests implicit incremental syntactic inferences underlie next-word predictions in autoregressive neural language models. \blfootnote{Code and materials: \url{https://github.com/eisape/incremental_parse_probe}}
\vspace{-1mm}
\end{abstract}

\section{Introduction}\vspace{-1mm}
The behavior of large-scale autoregressive neural language models (ALMs) appears to demonstrate impressive command of syntax \cite{Wilcox2019-lo, Hu2020-ml, Futrell2019-ow,  Gotlieb_Wilcox2021-qx,Arehalli2020-pb, Warstadt2020-js}. To what extent can we attribute this behavior to a model's maintaining and updating representations of incremental syntactic structures? 

Interpretability work on \emph{bidirectional} masked language models suggests that neural models of language may learn to encode syntactic structure through the geometry of their word embedding space. For example, \citet{Hewitt2019-yb} demonstrate that the dependency parse tree  of a sentence can be decoded by finding the minimum spanning tree  on pairwise syntactic distances regressed from linearly transformed contextualized word embeddings. 

Unlike the models considered in \citet{Hewitt2019-yb}, ALMs are \emph{unidirectional}, and the decision problem of incrementally deriving global syntactic structures has several nuances not present in the bidirectional setting. Sentence prefixes (e.g. ``I watched her duck ...'') can be ambiguous in their intended meaning in ways that are  disambiguated by their suffixes (e.g. ``quack'' vs. ``under the table''). Thus, incremental processors must maintain a belief state of parses that can be flexibly updated on the basis of future input. Insofar as language in the real world (e.g., speech, text, sign) is processed sequentially,  incremental disambiguation of structure and meaning is a task humans solve in everyday cognition, seemingly effortlessly \cite{Jurafsky1996-nz,Hale2001-ds, Levy2008-yw}. 

ALMs have been shown to recapitulate crucial features of human sentence processing \cite{Futrell2019-ow} and their representations moreover have been shown to align with  language processing in the brain better than their  bidirectional counterparts \cite{Schrimpf2021-gy,caucheteux2022brains}. We hypothesize that ALMs  learn and maintain correlates of incremental syntactic structures which play an important role in mediating model behavior. We investigate this hypothesis through the lens of \emph{counterfactual probing}, i.e., by learning classifiers over hidden states  of pretrained ALMs to predict linguistic properties, and then using the probes to intervene on model representations.  

We present a suite of probing architectures for decoding belief states of incremental structures  from pretrained Transformer ALMs given the model's hidden state, each of which embodies a different hypothesis for how incremental parse states might be encoded. We validate (and adjudicate between) our probes by using them to predict and control model behavior. Our results suggest that ALMs, through pretraining at scale, learn stack-like representations of syntax and use them in interpretable and controllable ways.

\begin{figure*}[t!]
	\centering
	\vspace{-3mm}
    \includegraphics[width=\linewidth, trim={2cm 8cm 5cm 7cm},clip]{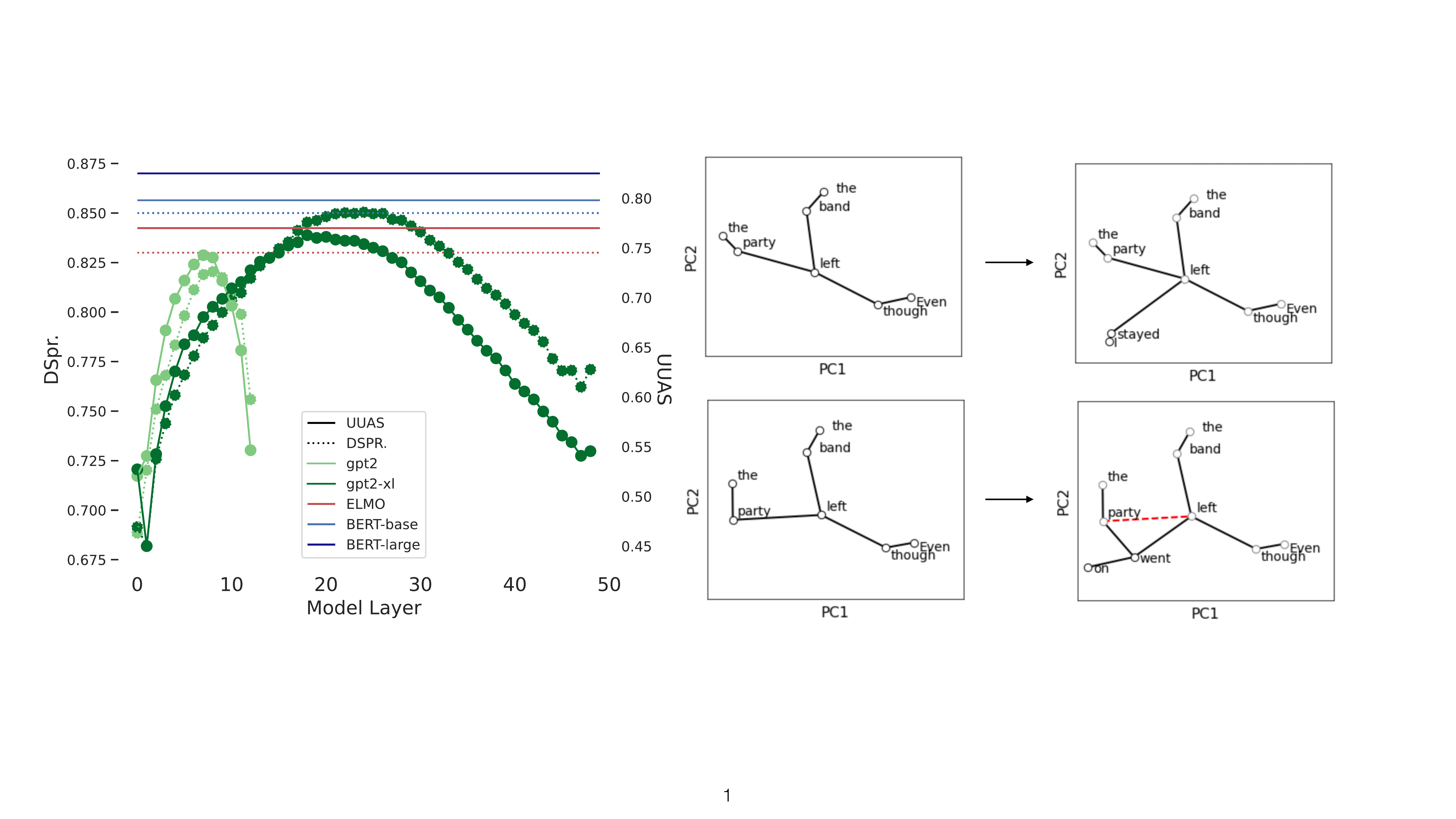}
    \vspace{-7mm}
    \caption{(Left) Results of applying the structural probe of \citet{Hewitt2019-yb} to GPT2. Displayed are the performances of the distance as quantified in both unlabeled undirected attachment score (UUAS) and Spearman correlation (Dspr.)  across layers. Layer 0 denotes uncontextualized embeddings. The best performing BERT and ELMo layers as reported in \citet{Hewitt2019-yb} are shown as horizontal lines. (Right) Dependency graphs  elicited from GPT2 small for the sentences in Figure~\ref{fig:npz_sents} via \citet{Hewitt2019-yb}'s structural probe before (left) and after (right) disambiguation, and then decoding with minimum spanning tree. Visualization is done by projecting the predicted pairwise distances into two dimensions via PCA. In the zero-complement condition (bottom), the ``went'' embedding is emitted between an existent dependency resulting in a new parse, effectively `deleting'  the  arc in red.}
	\label{fig:stuct_probe}
	\vspace{-4mm}
\end{figure*}

\section{Motivating Study}
Prior work has developed methods for extracting syntactic parses from neural language models given global context. In particular, \citet{Hewitt2019-yb} learn a distance function parameterized by a linear transformation ($B$) of the word embedding space of BERT,
\begin{align*}
\delta_{B}\left(\mathbf{h}_{i}^{}, \mathbf{h}_{j}^{}\right)^{2}=\left(B\left(\mathbf{h}_{i}^{}-\mathbf{h}_{j}^{}\right)\right)^{T}\left(B\left(\mathbf{h}_{i}^{}-\mathbf{h}_{j}^{}\right)\right),
\end{align*}
such that this distance approximates the distance of words $w_i, w_j$ in a dependency tree. After learning $B$ via regression, i.e., 
\begin{align*}
    \min _{B} \sum_{\ell} \frac{1}{\left|s^{\ell}\right|^{2}} \sum_{i, j}\left|\delta_{T^{\ell}}\left(w_{i}^{\ell}, w_{j}^{\ell}\right)-\delta_{B}\left(\mathbf{h}_{i}^{\ell}, \mathbf{h}_{j}^{\ell}\right)^{2}\right|,
\end{align*}
(here $\ell$ indexes the sentences in the training set, $|s^\ell|$ is the sentence length, and $\delta_{T^\ell}(w_i, w_j)$ is the tree distance between $w_i$ and $w_j$), 
the authors find that the minimum spanning tree obtained from $\delta_B(\cdot, \cdot)$ well-approximates the gold dependency tree in many cases, indicating that the geometry of contextualized word embeddings captures aspects of syntax.

\begin{figure}[t]
	\centering
    \includegraphics[scale=0.25]{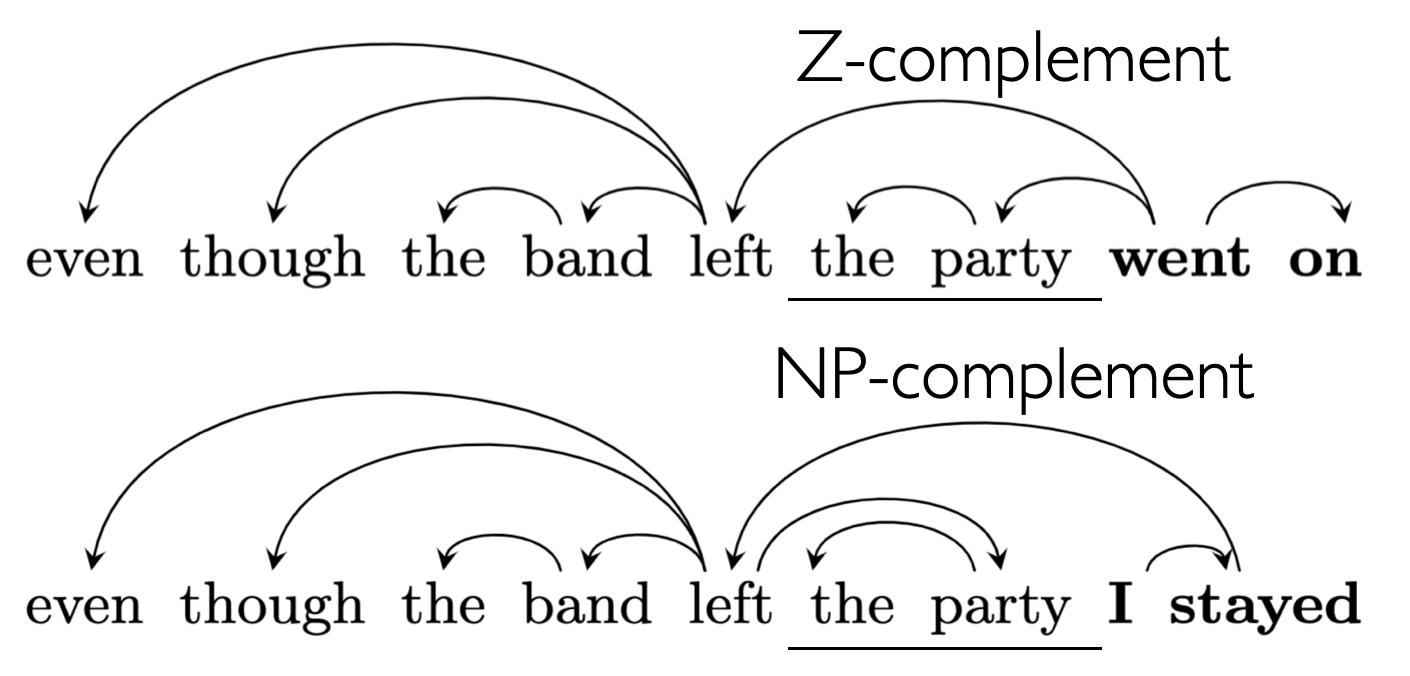}
    \vspace{-2mm}
    \caption{NP/Z ambiguous sentences and their  parses.}
     \vspace{-6mm}
	\label{fig:npz_sents}
\end{figure}
In an initial study, we run  \citet{Hewitt2019-yb}'s structural probe on the hidden states of GPT2 \cite{Radford2019-zt}, a unidirectional ALM, and find that GPT2's hidden states recover gold tree structures almost as well as similarly-sized bidirectional models (Figure~\ref{fig:stuct_probe}). For example GPT2-XL matches BERT-base in correlation with syntactic distances (Dspr.; .85 for both models).  We further observe that this probe can recover the correct structures even on sentences with ambiguous prefixes, such as the `NP/Z' ambiguity shown in Figure~\ref{fig:npz_sents}.
In this ambiguity, the shared prefix (``even though the band left the party'') is consistent with at least two interpretations: one in which the underlined span (``the party'') is a direct object noun phrase (NP) of the verb ``left'', and one in which the underlined content is the subject of a new noun phrase and ``left'' has a zero (Z) complement. This is a well-studied phenomenon in psycholinguistics, and humans make rapid and accurate inferences  in this setting \cite{BeverThomasG2013Tcbf}.

It is not at all obvious how a strictly left-to-right model,  whose representations of the words in context are \emph{static} (i.e., not affected by future words), can maintain a prefix representation that are \emph{dynamic} enough to encode global syntactic distances with high accuracy. We visualize the evolution of pairwise syntactic distances for the two sentences in Figure~\ref{fig:npz_sents} along with the minimum spanning trees in Figure~\ref{fig:stuct_probe} (right). We observe that parses probed from sentence prefixes up until the disambiguators are consistent with the NP-complement parse (i.e., where ``party'' is attached to ``left''), which is in line with previous work that analyzes GPT2's behavior on the NP/Z ambiguity \cite{Futrell2019-ow}. After observing ``went,'' GPT2 emits a representation whose distance is interleaved between ``left'' and  ``party'' (i.e., $\delta_B(\mathbf{h}_\text{party}, \mathbf{h}_\text{went}) < \delta_B(\mathbf{h}_\text{party},\mathbf{h}_\text{left})$), thus altering the minimum spanning tree to be consistent with the zero-complement parse. 

This analysis provides an initial hypothesis for how incremental parsing arises in ALMs from the perspective of syntactic distance---ALMs emit word representations that maintain syntactic uncertainty that can be exploited by later emissions, effectively editing inferred dependencies and bypassing the need to re-position past emissions. However, this analysis also reveals several limitations that preclude it as a viable model of incremental parsing in ALMs: (1) this method assumes a \textit{spanning-tree} and therefore cannot represent parses with open nodes, and (2) GPT2's behavior is consistent with a {probabilistic} parallel parser as it seems to entertain both possible parses until seeing the disambiguating words, but such a pure distance-based probe is not inherently probabilistic.\footnote{Though it is possible to derive a probabilistic parser by interpreting the summed distances as the energy of a globally normalized model.} In the following section, we develop several parameterizations of a probabilistic  incremental parser  that can represent incomplete syntactic structures.

\vspace{-1mm}
\section{An Incremental-Parse Probe for Autoregressive Language Models}
\vspace{-1mm}
We argue that the  stack representation of shift-reduce parsers provides a natural way to represent incomplete tree structures. In this paper, we work with the generative  arc-standard dependency formalism \cite{noauthor_undated-nr}, which maintains a {stack} of generated subtrees
$
  S = [s_1, s_2, s_3, ...]
$
(where the root of subtree $s_i$ is a word), and makes one of the following actions at each time step:
\begin{enumerate}
\vspace{-1.5mm}
\item[] \verb|LEFT-ARC|: pop the top two nodes,  create a new subtree by adding an arc $s_{1} \rightarrow s_{2}$, and push the new subtree onto the stack, 
\vspace{-1.5mm}
\item[] \verb|RIGHT-ARC|: pop the top two nodes, create a new subtree by adding an arc $s_{2} \rightarrow s_{1}$, and push the new subtree onto the stack, 
\vspace{-1.5mm}
\item[] \verb|GEN|: generate the next token.
\end{enumerate}
\begin{figure*}[t]
	\centering
	\vspace{-5mm}
    \includegraphics[scale=0.4, trim={2cm 12cm 26cm 12cm},clip]{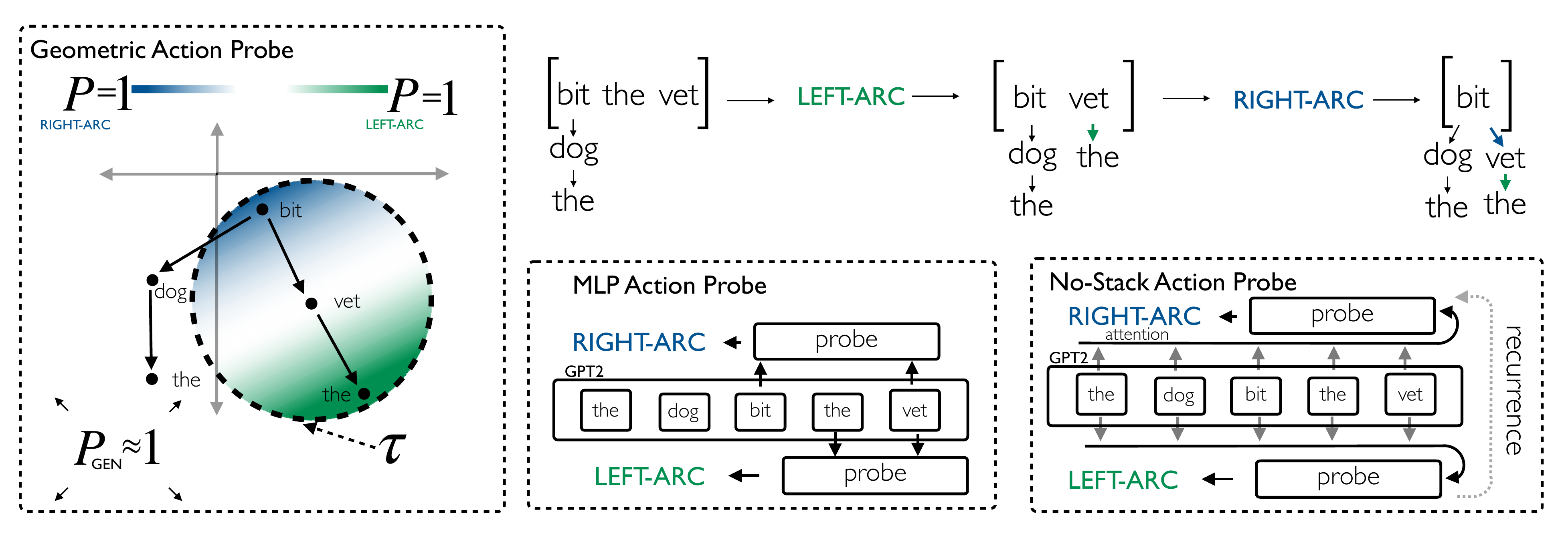}
    \vspace{-4mm}
    \caption{Schematics of each of the incremental parse probes parsing the sentence ``the dog bit the vet''. (Top) Incremental parse states (i.e., stacks of subtrees rooted by head words) are shown between the action transitions (\texttt{LEFT-ARC}, \texttt{RIGHT-ARC}) that connect the parse states. (Dotted) Visualizations of how each probe decides on the next actions. (Left) Geometric action probe (GAP) links action probabilities with its learned distance function $\delta_B$. The distance from the top node on the stack (``vet'' and ``the'' in the \texttt{LEFT-ARC} case) affects whether an arc is chosen. If the distance between the top two stack nodes is above a threshold $\tau$, then then we \texttt{GEN} with increasing probability. Relative depth (blue = shallow relative to ``vet'', green = deeper) predicts arc directionality probability. (Center) MLP Action probe (MAP) directly classifies each action by running the contextualized word embeddings for the top two nodes on the stack through an MLP. (Right) No-Stack Action Probe (NAP) classifies by attending over the entire prefix without maintaining an explicit stack.}
	\label{fig:architectures}
	\vspace{-4mm}
\end{figure*}

\noindent Generation starts with only \verb|ROOT| (an embedding learned independently for each probe) on the stack, $S =  [\verb|ROOT|]$, and terminates when only  \verb|ROOT| is left on the stack and the period token has been generated. Note that an action sequence $\mathbf{a}$ fully specifies a projective dependency structure.\footnote{Because ArcStandard can only recognize projective dependency trees, we exclude non-projective structures from training and evaluation in the sections to follow.} Hence, letting $\mathbf{a}_{\le n(t)}$ be the sequence of actions  up to (and including) the generation of $w_t$, we  use $\mathbf{a}_{\le n(t)}$ to represent the incremental (i.e., incomplete) tree structure state after emitting $w_t$.\footnote{We omit the generation of the next word (from GPT2's next word distribution) after predicting \texttt{GEN}, as we assume that sentences are given for probing purposes.} We design several probes that place distributions over action trajectories -   $P\left(\mathbf{a}_{\leq n\left(w_t\right)} \, | \, w_{<t}\right)$ given ALM embeddings $\mathbf{h}_{<t}$.

\subsection{Probe architectures}

We explore three probing architectures for parameterizing the action probabilities, each of which embodies a different hypothesis for how incremental parse states might be represented in ALMs.

\paragraph{Geometric Action Probe (GAP).} Our first architecture leverages the geometry of the embeddings and links the syntactic distances and depths derived from  \citet{Hewitt2019-yb} to action probabilities (Figure~\ref{fig:architectures}, left). The action probabilities are parameterized as below (where for brevity we omit the conditioning variables $w_{<t}$ and the previously generated actions):
\begin{align*}
   & P\left(\texttt{GEN}\right) =  \sigma\left(\frac{\delta_B\left(\mathbf{h}_{s_1},\mathbf{h}_{s_2}\right) -\tau}{\beta}\right),\\
    &P\left(\texttt{LEFT-ARC}\right) =  \left(1 - P\left(\texttt{GEN}\right)\right) \times \\ &\hspace{30mm} \sigma\left(\frac{\|\mathbf{h}_{s_1}\|_B - \|\mathbf{h}_{s_2}\|_B}{\beta}\right), \\
    & P\left(\texttt{RIGHT-ARC}\right) = \left(1 - P\left(\texttt{GEN}\right)\right) \times \\ &\hspace{30mm}  \sigma\left(\frac{\|\mathbf{h}_{s_2}\|_B - \|\mathbf{h}_{s_1}\|_B}{\beta}\right) ,
\end{align*}
\noindent where $\tau$ is a threshold parameter, $\beta$ is a temperature term, $\sigma(\cdot)$ is the sigmoid to turn distances into probabilities, and $\mathbf{h}_{s_i}$ is the ALM's representation for the root word of subtree $s_i$.  Note that the probability of \texttt{GEN} increases as the predicted syntactic distance between (the root words of) $s_1$ and $s_2$ increases, which intuitively captures the notion that there is less likely to be a link between $s_1$ and $s_2$ if their predicted distance is large.
The action probabilities for \texttt{LEFT-ARC} and \texttt{RIGHT-ARC} are based on comparing the predicted \emph{depths}, which we define to be the norm of the word embedding after a linear transformation as in \citet{Hewitt2019-yb}. For example, the probability of the arc $s_1 \rightarrow s_2$ increases as $s_2$ is predicted to be further away from the root than $s_1$ (i.e., relatively deeper nodes are the dependants of shallower nodes; see \figref{fig:architectures}). We initialize $B$ by pretraining on the distance and depth regression task from \citet{Hewitt2019-yb}. Because we also use a linear projection our probe can also be interpreted as a learned distance function on word representations.

\paragraph{MLP Action Probe (MAP).} The geometric action probe makes strong  assumptions about the underlying geometry of the representation space, i.e., that syntactic distances and depths are well-captured by linear transformations of the ALM's representations and that these measures can moreover be monotonically transformed to action probabilities via a sigmoid link function. This next variant relaxes this assumption and replaces the distance and depth-based linking function in GAP with a learned multilayer perceptron,
\begin{align*}
    P\left(a\right) \propto& \exp\left(e_{a}^{\top} \text{MLP}\left([\mathbf{h}_{s_1},\mathbf{h}_{s_2}]\right) + b_a \right)
\end{align*}
\noindent where $a \in \{\texttt{GEN},\texttt{LEFT-ARC},\texttt{RIGHT-ARC}\}$. As in the geometric version, we still make use of an explicit stack, but this variant  allows for an arbitrary link function from features of the model's representations to action probabilities.

\begin{figure*}[t]
\centering
\begin{minipage}{1\linewidth}
\removelatexerror
\begin{algorithm*}[H]
\SetAlgoLined
\footnotesize
\DontPrintSemicolon
\caption{\footnotesize Probe-Based Word-Synchronous Beam Search}\label{alg:three}
\KwIn{$\mathbf{w}$ \text{: Words in sentence.}}
\nonl \myinput{$P_{\text{probe}}$: \text{Incremental parse probe.}}
\nonl \myinput{$k_{\text{action}}$: \text{Number of action n-grams to consider between words.}}
\nonl \myinput{$k_{\text{word}}$: \text{Max number of parses to consider for each word.}}
\nonl \myinput{$k_{\text{out}}$: \text{Number of parses to output.}}
\KwOut{$\mathbf{B}^{\text{\upshape out}}$\text{: beam of terminal action sequences}}
\nonl \;
$\mathbf{B}, \mathbf{B}^{\text{\upshape out}} \gets [[\texttt{ROOT}]], [\,]$ \; 
\While{$|\mathbf{B}^{\text{\upshape out}}| \leq k_{\text{\upshape out}}$}{
\For{$n \in [1, ..., |\mathbf{w}|]$}{
$\mathbf{B}^{\text{\upshape word}} \gets [\,]$ \;
  $\textbf{h}_{<n} \gets \text{GPT2}\left(\textbf{w}_{<n}\right)$ \tcp*{get hidden states from GPT2} 
\While{$\mathbf{B}$ \upshape not empty}{

 $\mathbf{a} \gets \text{pop}\left(\mathbf{B}\right)$\;
$\{\mathbf{a}^{1},\mathbf{a}^{2}, ..., \mathbf{a}^{k_{\text{\upshape action}}}\} \gets  \text{BeamSearch}\left(P_{\text{probe}}, \mathbf{a}, \textbf{h}_{<n}, k_{\text{\upshape action}}\right)$\tcp*{search until GEN or termination} 

\For{$\mathbf{a}_{\text{\upshape new}} \in \{\mathbf{a}^{1},\mathbf{a}^{2}, ..., \mathbf{a}^{k_{\text{\upshape action}}}\}$}{
\lIf{$n = |\mathbf{w}|$:}{
$\text{push}\left(\mathbf{B}^{\text{\upshape out}}, \mathbf{a}.\text{\upshape append}\left(\mathbf{a}_{\text{\upshape new}}\right)\right)$} 

\lElse{$\text{: push}\left(\mathbf{B}^{\text{\upshape word}}, \mathbf{a}.\text{\upshape append}\left(\mathbf{a}_{\text{\upshape new}}\right)\right)$}
  }
  }
    $\mathbf{B} \gets \text{top-k}\left(\mathbf{B}^{\text{\upshape word}}, k_{\text{\upshape word}}\right)$\tcp*{synchronize beam at word-level} 
  }
  }
\Return $\mathbf{B}^{\text{\upshape out}}$\tcp*{return beam with complete parse trees} 
  \label{alg:beamsearch}
\end{algorithm*}
\end{minipage}
\vspace{-3mm}
  \caption*{Algorithm 1: Probe-based word-synchronous beam search. We use beam search (line 8) as a subroutine to extend action sequences in the beam with action n-gram continuations. After the generation of each word the beam is pruned to the top-$k_{\text{\upshape word}}$ best parses (line 12).}
  \vspace{-3mm}
\end{figure*}

\paragraph{No-Stack Action Probe (NAP).} Our final variant relaxes assumptions about both the linking function and the existence of an explicit stack. This approach, which is closely related to \citet{Qian2021-ez}, simply predicts the sequence of actions \emph{between} two words $w_t$ and $w_{t+1}$ using the action history and the hidden representations $\mathbf{h}_{<t}$,
\begin{align*}
    & \mathbf{v}_{j} = \text{Action-LSTM}(\mathbf{v}_{j-1}, a_{j}), \\
    & \tilde{\mathbf{h}} = \text{Attention}(\mathbf{h}_{<t}, \mathbf{v}_j), \\
    &P\left(a\right) \propto \exp\left(e_{a}^{\top} \text{MLP}\left([\tilde{\mathbf{h}}, \mathbf{v}_j]\right) + b_a \right).
\end{align*}
Here the action LSTM's hidden state $\mathbf{v}_j$ is used to attend over the previous word representations $\mathbf{h}_{<t}$ to obtain a context vector $\tilde{\mathbf{h}}$, which is combined with the hidden state to produce a distribution over the following action.

We train each of these architectures\footnote{Due to resource constraints we only train NAP on GPT-2 small.} on the ground truth action trajectories in the Penn Treebank (PTB) \cite{Marcus1993-nc} and use \texttt{gpt2} and \texttt{gpt2-xl} model checkpoints from HuggingFace \cite{Wolf2019-vb} as our ALM text encoders. All hyperparameters are reported in Appendix \ref{sec:app-training}.

\vspace{-1mm}
\section{Experiments}
\vspace{-1mm}
We evaluate each of our probes on their parsing performance over the PTB test split as well as their ability to predict and control model behavior.

\subsection{Parsing Performance}

\setlength{\textfloatsep}{1pt}
For parsing performance, we evaluate with action perplexity (PPL) and unlabeled attachment score on the PTB test set. For the latter, we extend the word-synchronous beam search algorithm for decoding from generative neural parsers \cite{Stern2017-vm}  to the incremental probe setting. 
\begin{figure*}[t!]
	\centering
	    \vspace{-6mm}
    \includegraphics[trim={3.58cm 9.7cm 5.cm 9cm},clip,scale=0.27]{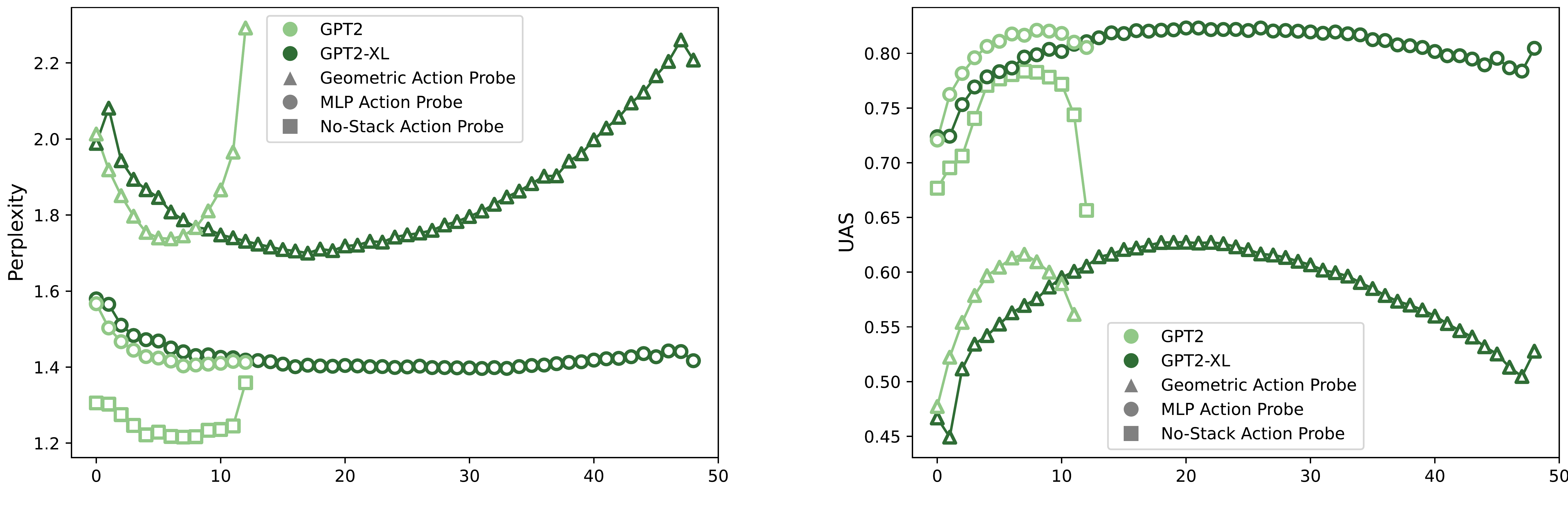}
    \vspace{-9mm}
    \caption{Action perplexity (left) and UAS (right) for each of the incremental-parse probes.}
	\label{fig:probe_eval}
	    \vspace{-4mm} 
\end{figure*}
This \textit{probe-based} word-synchronous beam search algorithm uses an incremental parse probe to decode action sequences \textit{between} word emissions with beam search. Because the word emissions from the ALMs we consider are contextualized, this quantity implicitly conditions on the previous action sequence including all of the words generated in the prefix so far\footnote{That is, even though our incremental-parse probes only predict $\texttt{GEN}$, action likelihoods in our probes are conditioned on $\texttt{GEN}(w)$ for words in the prefix: $w \in w_{\leq t}$.} thus allowing us to predict the  likelihood:$\displaystyle \prod_{a \in \pi\backslash \textbf{w}} P(a)$ (where $\pi\backslash \textbf{w}$ are the actions in any parse $\pi$ excluding the likelihoods of words conditioned on actions). 

This algorithm is an interesting testbed for ALMs because it relies \textit{only} on the syntactic disambiguation implicit in the ALMs hidden states to update its beam of parses (compared to models such as RNNGs \cite{Dyer2016-px} whose next-word distribution is explicitly conditioned on the stack states). This decoding scheme is shown in Algorithm 1 where $k_{\text{action}}$ is the  the number of action n-grams to consider between model emissions, and $k_{\text{word}}$ is number of parses to consider at word boundaries. During decoding we only consider the valid set of actions for a given parse state given the rules of ArcStandard. 

The results of these experiments are shown in Figure \ref{fig:probe_eval}, where we set $k_{\text{action}} = k_{\text{word}} = 10$. MAP outperforms GAP as expected (since it makes weaker assumptions). Interestingly, while NAP outperforms the other probes in terms of PPL, when evaluated as an incremental parser, it has worse UAS than MAP. This is significant because while MAP makes explicit use of a stack (i.e., the embeddings it uses to make decisions are entirely determined by the stack), NAP does not and instead uses the attention distribution over the prefix at each time step to implicitly encode the stack. We take these results as an important data point in adjudicating between representational hypotheses of incremental syntax in ALMs. Under the assumption that high-performance ALMs are adept incremental parsers---as has been previously demonstrated \cite{Marvin2018-oc, Hu2020-ml}---the high performance of MAP suggests our best current model of syntactic parsing in ALMs may be stack-based but not necessarily geometry-based. 

Probe performance to an extent implies the \emph{existence} of implicit representaitons of syntax with ALMs. However, it does not necessarily imply that these structures mediate model behavior.  In the following section, we conduct experiments to see whether we can predict and control model behavior with our probes.

\vspace{-1mm}
\subsection{Probing Incremental Disambiguation}
\vspace{-1mm}
We evaluate our probes to predict model behavior. For this purpose, we extend the dataset of NP/Z ambiguous sentences of \citet{Futrell2019-ow} by  augmenting it to include continuations that disambiguate toward the Zero- and NP-complement interpretations (Figure \ref{fig:npz_sents} shows an example data point). We also include continuations that are consistent with both parses (``Even though the band left the party \textit{that was raging} [...]'') or neither parse (we use the period token, which is ungrammatical in either case, i.e., ``Even though the band left the party \textit{.}'') for a more complete comparison of the effects of disambiguation. Concretely, the likelihoods of `Both' and `Neither' continuations are expected to increase and stay the same, respectively, regardless of the direction of disambiguation.

\begin{figure}[t!]
	\centering
    \includegraphics[trim={19cm 4.8cm 18cm 7cm},clip,scale=0.3]{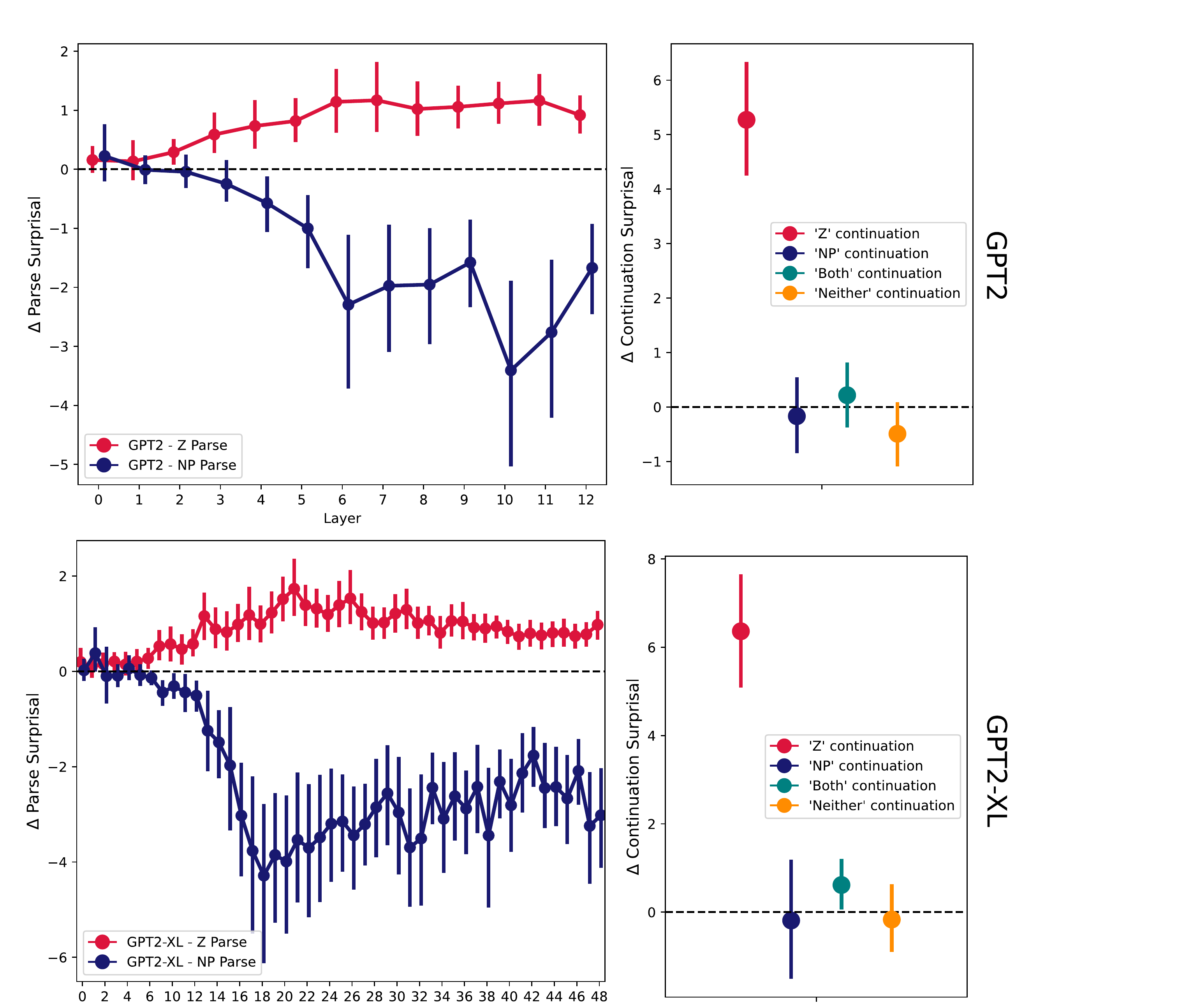}
    \caption{Disambiguation as evaluated by MAP and model behavior. (Right) Difference in \emph{GPT2} surprisal over the continuations in our corpus from the unambiguous condition to the ambiguous condition (higher means the model is more surprised in the unambiguous condition). (Left) Difference in (MAP) \emph{probe} surprisal over the disambiguating parse action. Error bars show 95\% confidence intervals across our corpus.}
	\label{fig:behaviour}
\end{figure}

\paragraph{Predicting Behavior.} First, we replicate \citet{Futrell2019-ow}'s result showing that ALMs are sensitive to verb transitivity and use it to guide next-word predictions, resulting in incremental parser-like behavior. Specifically, we show that  changing the verb in NP/Z-ambiguous prefixes from intransitive (unambiguously Z-complement favoring) to transitive (ambiguous), e.g.,:

\ex.\label{verbtrans} Even though the band \textbf{left} [ambiguous],
    Even though the band \textbf{performed} [unambiguous],

causes ALMs to prefer the continuation consistent with the zero-complement parse. Following \citet{Futrell2019-ow} we use the ALM's surprisal over the words in the continuation\footnote{Note that as the sentence is not fully disambiguated until the period is reached. We include the period token in our surprisal estimates.} to quantify the ALM's expectations, i.e.,  $S(\text{``went on.''}) =  -\log P_\text{model}(\text{``went on.''} | \textit{prefix})$. In Figure \ref{fig:behaviour} we show the change (difference) in surprisal from the unambiguous to the ambiguous case and observe that GPT2 and GPT2-XL both find the Z continuation more surprising in the presence of a transitive verb (Figure~\ref{fig:behaviour}, right). Somewhat surprisingly, the surprisal of the NP continuation is unchanged by verb transitivity. As expected, the surprisal of the `Both' and `Neither' conditions are largely unchanged between conditions.

We find that our probes can predict model behavior in this setting. We target the first action that differentiates the NP and Z parses, which corresponds to the decision to place and arc from the verb to the head of the noun phrase (Figure~\ref{fig:npz_sents}) and plot the difference in surprisal of this decision from the intransitive condition to the transitive condition (Figure~\ref{fig:behaviour}, left). This probe illustrates a preference for the Z-complement parse in the presence of an intransitive verb which provides representational evidence for incremental parse disambiguation on the part of the model.

\begin{figure}[t!]
	\centering
    \includegraphics[trim={25cm 2.3cm 25cm 2.3cm},clip,scale=0.37]{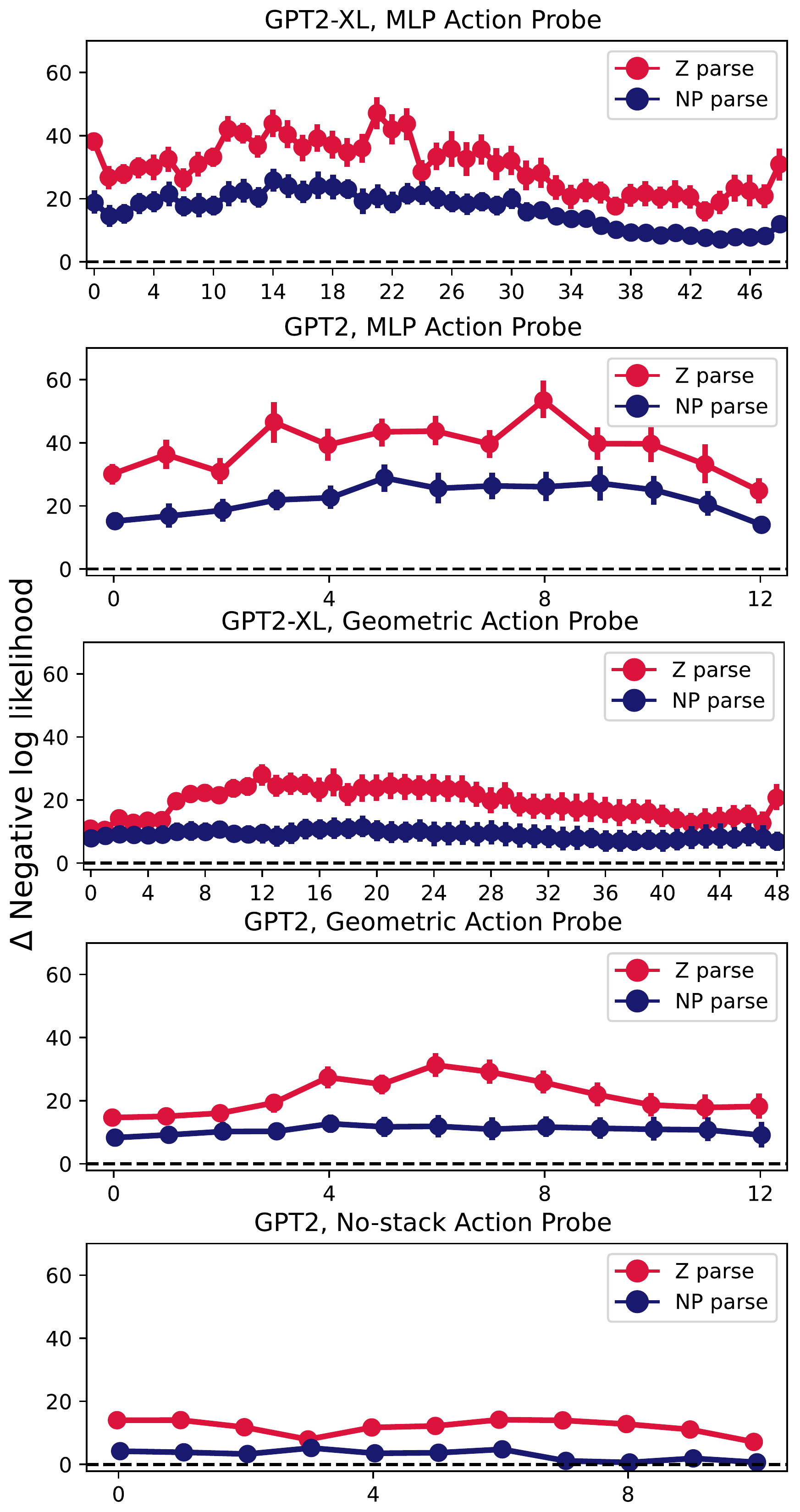}
        \vspace{-2mm}
    \caption{Effect of sentence continuations on parse likelihood. For each of the incremental parse probes, we produce the negative log likelihood for each parse given 1) congruent suffix words and 2) incongruent suffix words (larger is better) and show each probe assigns higher negative log-likelihood to parses when matched with an incongruent suffix (lines above origin in all cases).}
	\label{fig:behaviour2}
\end{figure}

In a further study, we present each of the probes with the two NP/Z parses in two conditions: 1) where the sentence suffix was consistent with the parse being probed, and 2) where the sentence suffix was consistent with the other parse. This can be seen as a probabilistic version of Figure~\ref{fig:stuct_probe} (right). We plot the difference in negative log-likelihoods from condition 2 to condition 1 in Figure~\ref{fig:behaviour2}. As expected, we find each of the probes assign higher negative log likelihood (i.e., lower likelihood) to parses when matched with incongruent suffixes. This  effect is especially pronounced for the MAP probe.

\begin{figure*}[t!]
	\centering
    \includegraphics[width=\linewidth, trim={18cm 15cm 20.8cm 11.3cm},clip]{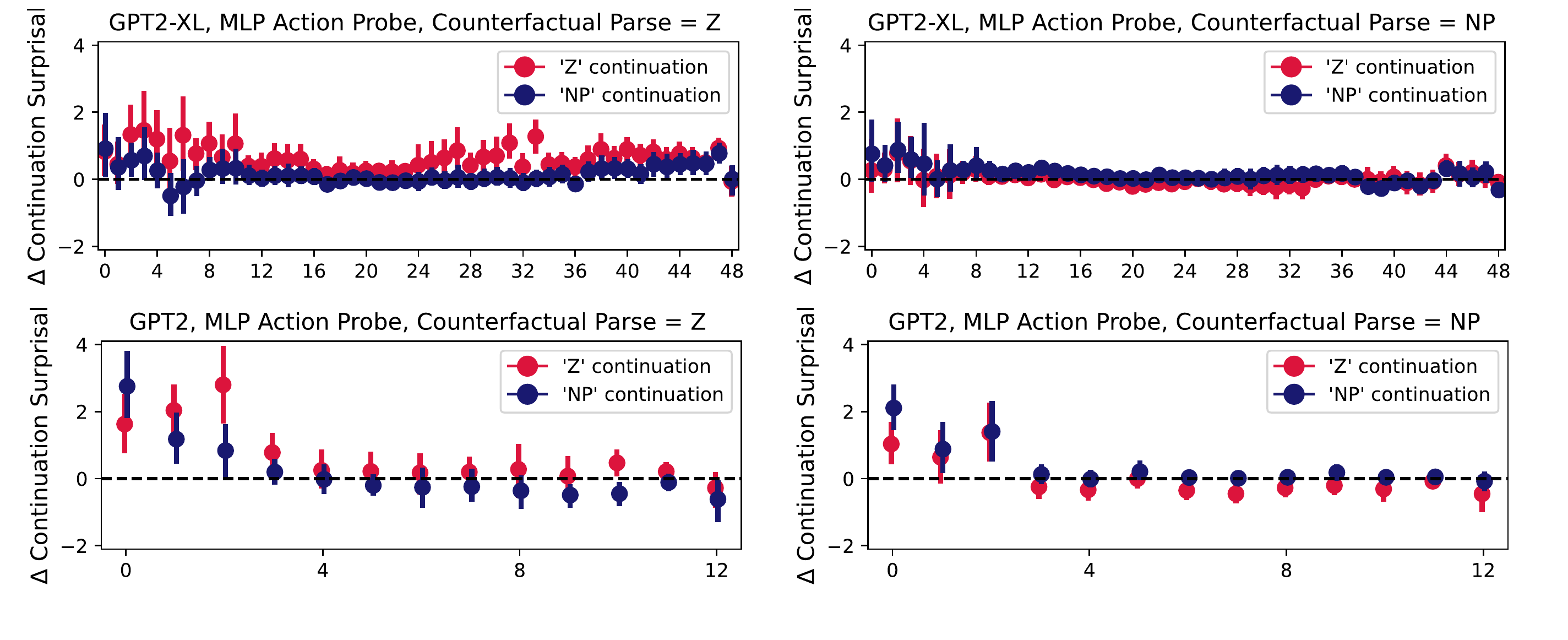}
    	\vspace{-11mm}
    \caption{Results of counterfactual syntactic perturbations. The y-axis shows the surprisal difference over the sentence continuations from unperturbed GPT2 to GPT2 perturbed  with counterfactual embeddings. We see differences in the expected directions for the Z-complement and NP-complement (i.e., red above blue in the plots on the left and blue above red in the rightward plots) completions for MAP across model. Error bars show 95\% confidence intervals across our corpus.}
	\label{fig:cfx}
	\vspace{-5mm}
\end{figure*}

\paragraph{Editing Parse States.} Finally, we use counterfactual analysis \cite{Tucker2021-sj, Tucker2022-yd} at the level of ALM hidden states to probe the extent to which our probes capture essential information about incremental parses that mediate model behavior. Our approach generates counterfactual ALM hidden states by propagating the loss of a particular parse state ($\mathbf{a}$) from our probe output to model embeddings. Specifically, we iteratively update GPT2 hidden states with the following: $\widehat{\mathbf{h}} = \mathbf{h} +  \epsilon \nabla_{ \mathbf{h}}  P_{\text{probe}}\left(\mathbf{a} \mid \mathbf{h}  \right)$, for 
 a small step size $\epsilon$ (see Appendix \ref{fig:cfx} for further details). Crucially, we perturb towards incremental, incomplete structures before a word is generated, rather than complete structures.
 
 We conjecture that if our probes pick up on essential syntactic information in model embeddings, they should be able to generate model embeddings that produce the expected effect on model behaviour (e.g., perturbing toward the Z-complement parse should make the Z-complement continuation more likely). We generate two sets of counterfactual embeddings (one for each syntactic interpretation; Figure~\ref{fig:npz_sents}) for each sentence in our modified NP/Z data set. We do this for each layer of the models considered for each of our probes and evaluate the effect of these treatments by generating surprisal estimates from each of the counterfactual embeddings.\\
 \indent The results of our analysis are shown in Figure \ref{fig:cfx}. We find that counterfactuals generated with MAP produce the predicted effect on both GPT2 and GPT2-XL for the Z- and NP-congruent continuations (but not for the `Both' or `Neither' conditions; Appendix~\ref{sec:app-cfx}). This effect is significant across several layers of both models and is most significant at layer 2 of GPT2 small, where Z-complement-congruent continuations increase in likelihood by $\sim$8 fold (3 nats) on average after model intervention. We find that counterfactuals generated with the other probes do not mediate next-word predictions in the expected way (Appendix \ref{sec:app-cfx}).

\section{Related Work}
Our work is related to the growing literature quantifying the syntactic sensitivity of neural models of language at  behavioral \cite{Linzen2021-mh,Aina2021-wh, Lakretz2021-tj} and  representational  \cite{Pimentel2020-jl1,Hewitt2019-yb,Manning2020-vp,Muller-Eberstein2022-ly} levels. Of particular note among these studies is the dependency-arc labeling task introduced in \citet{Tenney2019-al} which bears resemblance to our MAP architecture. Our work extends this literature by developing several new types of syntactic probes in the incremental setting. 

Similar to our work but in semantics, \citet{Li2021-gx} use probes to interpret text-encoders as maintaining an evolving semantic information state while interacting with a text-based game. Relatedly, our work is among others that attempt to control neural models of language for counterfactual analysis  or controllable AI more generally \cite{De_Cao2021-dc, Ravfogel2021-jk, Elazar2020-fw, Meng2022-zx, Dathathri2019-md}. 

Lastly, we expect our work to be of consequence to computational psycholinguistics, where neural language models and incremental parsers have been historically prominent as candidate cognitive models \cite{Hale2018-df, Hale2001-ds,Wilcox2020-uo, Levy2008-yw,Jurafsky1996-nz, Eisape2020-jh}. This work draws representational parallels between these two models.

\section{Conclusion}
Motivated by recent work showing human-like incremental parsing behaviour in ALMs, this work extends structural probing to the incremental setting. We find that autoregressive language models can perform on par with bidirectional models in terms of global parsing metrics despite the fact that an ALM's representation cannot condition on future words. We present several hypotheses of the representation of incremental structure in ALMs, instantiate them in probe architectures, and evaluate their explanatory power across a wide range of experiment types: parsing performance, predicting model behavior, and counterfactual analyses. Cumulatively, MAP performs the best, which is significant because this architecture incorporates strong constraints based on linguistic theory (i.e., stack-based representations) while still being relative agnostic (compared to the geometric probe) to the details of how parse probabilities might be encoded. This suggests that despite a lack of explicit feedback, language models not only rediscover linguistic structure through pretraining but learn to make `inferences' over this structure in real-time comprehension.

\section{Limitations}
The methods presented here have several limitations. Principally, the question of what constitutes a probe remains a potential confound \cite{Hewitt2019-uy, Belinkov2021-tq}. While we show several effects that suggest our probes identify meaningful syntactic information, including counterfactual perturbations, our counterfactual effects, though significant for many of the layers of the models considered, are small in absolute terms. Our approach to mitigating these has been to apply a varied set of analyses that, in aggregate, adjudicate between the hypotheses we present here. Finally, our probes rely on a particular oracle---it is possible that action sets from other oracles (e.g. ArcEager) are better-suited towards representing incremental syntactic structures than ArcStandard.

\section{Ethical and Broader Impacts}
Language models are increasingly used for tasks beyond language that assume the ability of the language model to structure their input and make `decisions' in real-time. Our study targets this ability in a basic linguistic phenomenon and thus has the potential to be useful for interpretability and alignment. While the model control aspects of our study pose some risks in that it may enable malicious parties to generate harmful content with language models, the authors believe the benefits in terms of interpretability reasonably balance these.

\section*{Acknowledgments}
We would like to thank Peng Qian, Mycal Tucker, Jacob Andreas, Josh Tenenbaum, and Ted Gibson for helpful discussions. TE acknowledges support from the GEM consortium and the National Science Foundation Graduate Research Fellowship under Grant No. 1745302. RPL acknowledges support from NSF award BCS-2121074, NIH award U01-NS121471, and a Newton Brain Science Award. YK acknowledges support from MIT-IBM Watson AI lab. Lastly, our codebase is built, in part, on code from open source projects for prepossessing and parsing PTB \cite{Qi2017-zu,Hewitt2019-yb,Noji2021-zu} as well deploying transformer language models \cite{Paszke_undated-ly,Wolf2019-vb}.

\bibliography{anthology,custom, tiwa}

\begin{thebibliography}{46}
\expandafter\ifx\csname natexlab\endcsname\relax\def\natexlab#1{#1}\fi

\bibitem[{Aina and Linzen(2021)}]{Aina2021-wh}
Laura Aina and Tal Linzen. 2021.
\newblock \href {http://arxiv.org/abs/2109.07848} {The language model
  understood the prompt was ambiguous: Probing syntactic uncertainty through
  generation}.

\bibitem[{Arehalli and Linzen(2020)}]{Arehalli2020-pb}
Suhas Arehalli and Tal Linzen. 2020.
\newblock Neural language models capture some, but not all, agreement
  attraction effects.

\bibitem[{Belinkov(2021)}]{Belinkov2021-tq}
Yonatan Belinkov. 2021.
\newblock Probing classifiers: Promises, shortcomings, and advances.

\bibitem[{Bever(1970)}]{BeverThomasG2013Tcbf}
Thomas~G Bever. 1970.
\newblock The cognitive basis for linguistic structures.
\newblock In \emph{Cognition and the development of language}. Wiley.

\bibitem[{Caucheteux and King(2022)}]{caucheteux2022brains}
Charlotte Caucheteux and Jean-R{\'e}mi King. 2022.
\newblock Brains and algorithms partially converge in natural language
  processing.
\newblock \emph{Communications Biology}, 5(1):1--10.

\bibitem[{Dathathri et~al.(2019)Dathathri, Madotto, Lan, Hung, Frank, Molino,
  Yosinski, and Liu}]{Dathathri2019-md}
Sumanth Dathathri, Andrea Madotto, Janice Lan, Jane Hung, Eric Frank, Piero
  Molino, Jason Yosinski, and Rosanne Liu. 2019.
\newblock \href {http://arxiv.org/abs/1912.02164} {Plug and play language
  models: A simple approach to controlled text generation}.

\bibitem[{De~Cao et~al.(2021)De~Cao, Schmid, Hupkes, and Titov}]{De_Cao2021-dc}
Nicola De~Cao, Leon Schmid, Dieuwke Hupkes, and Ivan Titov. 2021.
\newblock \href {http://arxiv.org/abs/2112.06837} {Sparse interventions in
  language models with differentiable masking}.

\bibitem[{Dyer et~al.(2016)Dyer, Kuncoro, Ballesteros, and Smith}]{Dyer2016-px}
Chris Dyer, Adhiguna Kuncoro, Miguel Ballesteros, and Noah~A Smith. 2016.
\newblock \href {http://arxiv.org/abs/1602.07776} {Recurrent neural network
  grammars}.

\bibitem[{Eisape et~al.(2020)Eisape, Zaslavsky, and Levy}]{Eisape2020-jh}
Tiwalayo Eisape, Noga Zaslavsky, and Roger Levy. 2020.
\newblock Cloze distillation: Improving neural language models with human
  {Next-Word} prediction.
\newblock In \emph{Proceedings of the 24th Conference on Computational Natural
  Language Learning}, pages 609--619, Online. Association for Computational
  Linguistics.

\bibitem[{Elazar et~al.(2020)Elazar, Ravfogel, Jacovi, and
  Goldberg}]{Elazar2020-fw}
Yanai Elazar, Shauli Ravfogel, Alon Jacovi, and Yoav Goldberg. 2020.
\newblock \href {http://arxiv.org/abs/2006.00995} {Amnesic probing: Behavioral
  explanation with amnesic counterfactuals}.

\bibitem[{Falcon et~al.(2019)}]{falcon2019pytorch}
William Falcon et~al. 2019.
\newblock Pytorch lightning.
\newblock \emph{GitHub. Note: https://github.
  com/PyTorchLightning/pytorch-lightning}, 3(6).

\bibitem[{Futrell et~al.(2019)Futrell, Wilcox, Morita, Qian, Ballesteros, and
  Levy}]{Futrell2019-ow}
Richard Futrell, Ethan Wilcox, Takashi Morita, Peng Qian, Miguel Ballesteros,
  and Roger Levy. 2019.
\newblock Neural language models as psycholinguistic subjects: Representations
  of syntactic state.
\newblock In \emph{Proceedings of the 2019 Conference of the North {A}merican
  Chapter of the Association for Computational Linguistics: Human Language
  Technologies, Volume 1 (Long and Short Papers)}, pages 32--42, Minneapolis,
  Minnesota. Association for Computational Linguistics.

\bibitem[{Hale(2001)}]{Hale2001-ds}
John Hale. 2001.
\newblock A probabilistic {E}arley parser as a psycholinguistic model.
\newblock In \emph{Second Meeting of the North {A}merican Chapter of the
  Association for Computational Linguistics}.

\bibitem[{Hale et~al.(2018)Hale, Dyer, Kuncoro, and Brennan}]{Hale2018-df}
John Hale, Chris Dyer, Adhiguna Kuncoro, and Jonathan~R Brennan. 2018.
\newblock \href {http://arxiv.org/abs/1806.04127} {Finding syntax in human
  encephalography with beam search}.

\bibitem[{Hewitt and Liang(2019)}]{Hewitt2019-uy}
John Hewitt and Percy Liang. 2019.
\newblock \href {http://arxiv.org/abs/1909.03368} {Designing and interpreting
  probes with control tasks}.

\bibitem[{Hewitt and Manning(2019)}]{Hewitt2019-yb}
John Hewitt and Christopher~D Manning. 2019.
\newblock A structural probe for finding syntax in word representations.
\newblock In \emph{Proceedings of the 2019 Conference of the North American
  Chapter of the Association for Computational Linguistics: Human Language
  Technologies, Volume 1 (Long and Short Papers)}, pages 4129--4138.

\bibitem[{Hu et~al.(2020)Hu, Gauthier, Qian, Wilcox, and Levy}]{Hu2020-ml}
Jennifer Hu, Jon Gauthier, Peng Qian, Ethan Wilcox, and Roger~P Levy. 2020.
\newblock \href {http://arxiv.org/abs/2005.03692} {A systematic assessment of
  syntactic generalization in neural language models}.

\bibitem[{Jurafsky(1996)}]{Jurafsky1996-nz}
Daniel Jurafsky. 1996.
\newblock A probabilistic model of lexical and syntactic access and
  disambiguation.
\newblock \emph{Cogn. Sci.}, 20(2):137--194.

\bibitem[{Kingma and Ba(2014)}]{Kingma2014-nr}
Diederik~P Kingma and Jimmy Ba. 2014.
\newblock \href {http://arxiv.org/abs/1412.6980} {Adam: A method for stochastic
  optimization}.

\bibitem[{Lakretz et~al.(2021)Lakretz, Desbordes, Hupkes, and
  Dehaene}]{Lakretz2021-tj}
Yair Lakretz, Th{\'e}o Desbordes, Dieuwke Hupkes, and Stanislas Dehaene. 2021.
\newblock \href {http://arxiv.org/abs/2110.07240} {Causal transformers perform
  below chance on recursive nested constructions, unlike humans}.

\bibitem[{Levy(2008)}]{Levy2008-yw}
Roger Levy. 2008.
\newblock Expectation-based syntactic comprehension.
\newblock \emph{Cognition}, 106(3):1126--1177.

\bibitem[{Li et~al.(2021)Li, Nye, and Andreas}]{Li2021-gx}
Belinda~Z Li, Maxwell Nye, and Jacob Andreas. 2021.
\newblock \href {http://arxiv.org/abs/2106.00737} {Implicit representations of
  meaning in neural language models}.

\bibitem[{Linzen and Baroni(2021)}]{Linzen2021-mh}
Tal Linzen and Marco Baroni. 2021.
\newblock Syntactic structure from deep learning.
\newblock \emph{Annu. Rev. Linguist.}, 7(1):195--212.

\bibitem[{Manning et~al.(2020)Manning, Clark, Hewitt, Khandelwal, and
  Levy}]{Manning2020-vp}
Christopher~D Manning, Kevin Clark, John Hewitt, Urvashi Khandelwal, and Omer
  Levy. 2020.
\newblock Emergent linguistic structure in artificial neural networks trained
  by self-supervision.
\newblock \emph{Proc. Natl. Acad. Sci. U. S. A.}

\bibitem[{Marcus et~al.(1993)Marcus, Santorini, and
  Marcinkiewicz}]{Marcus1993-nc}
Mitchell~P Marcus, Beatrice Santorini, and Mary~Ann Marcinkiewicz. 1993.
\newblock Building a large annotated corpus of {E}nglish: The {P}enn
  {T}reebank.
\newblock \emph{Comput. Linguist.}, 19(2):313--330.

\bibitem[{Marvin and Linzen(2018)}]{Marvin2018-oc}
Rebecca Marvin and Tal Linzen. 2018.
\newblock \href {http://arxiv.org/abs/1808.09031} {Targeted syntactic
  evaluation of language models}.

\bibitem[{Meng et~al.(2022)Meng, Bau, Andonian, and Belinkov}]{Meng2022-zx}
Kevin Meng, David Bau, Alex Andonian, and Yonatan Belinkov. 2022.
\newblock \href {http://arxiv.org/abs/2202.05262} {Locating and editing factual
  knowledge in {GPT}}.

\bibitem[{M{\"u}ller-Eberstein et~al.(2022)M{\"u}ller-Eberstein, van~der Goot,
  and Plank}]{Muller-Eberstein2022-ly}
Max M{\"u}ller-Eberstein, Rob van~der Goot, and Barbara Plank. 2022.
\newblock \href {http://arxiv.org/abs/2203.12971} {Probing for labeled
  dependency trees}.

\bibitem[{Nivre(2004)}]{noauthor_undated-nr}
Joakim Nivre. 2004.
\newblock Incrementality in deterministic dependency parsing.
\newblock pages 50--57.

\bibitem[{Noji and Oseki(2021)}]{Noji2021-zu}
Hiroshi Noji and Yohei Oseki. 2021.
\newblock \href {http://arxiv.org/abs/2105.14822} {Effective batching for
  recurrent neural network grammars}.

\bibitem[{Paszke et~al.()Paszke, Gross, and Chintala}]{Paszke_undated-ly}
Adam Paszke, Sam Gross, and Soumith Chintala.
\newblock Automatic differentiation in {PyTorch}.

\bibitem[{Pimentel et~al.(2020)Pimentel, Valvoda, Maudslay, Zmigrod, Williams,
  and Cotterell}]{Pimentel2020-jl1}
Tiago Pimentel, Josef Valvoda, Rowan~Hall Maudslay, Ran Zmigrod, Adina
  Williams, and Ryan Cotterell. 2020.
\newblock {Information-Theoretic} probing for linguistic structure.

\bibitem[{Qi and Manning(2017)}]{Qi2017-zu}
Peng Qi and Christopher~D Manning. 2017.
\newblock \href {http://arxiv.org/abs/1705.04434} {Arc-swift: A novel
  transition system for dependency parsing}.

\bibitem[{Qian et~al.(2021)Qian, Naseem, Levy, and
  Fernandez~Astudillo}]{Qian2021-ez}
Peng Qian, Tahira Naseem, Roger Levy, and Ram{\'o}n Fernandez~Astudillo. 2021.
\newblock Structural guidance for transformer language models.
\newblock In \emph{Proceedings of the 59th Annual Meeting of the Association
  for Computational Linguistics and the 11th International Joint Conference on
  Natural Language Processing (Volume 1: Long Papers)}, pages 3735--3745,
  Online. Association for Computational Linguistics.

\bibitem[{Radford et~al.(2019)Radford, Wu, Child, Luan, Amodei, and
  Sutskever}]{Radford2019-zt}
Alec Radford, Jeffrey Wu, Rewon Child, David Luan, Dario Amodei, and Ilya
  Sutskever. 2019.
\newblock Language models are unsupervised multitask learners.
\newblock \emph{OpenAI Blog}, 1(8):9.

\bibitem[{Ravfogel et~al.(2021)Ravfogel, Prasad, Linzen, and
  Goldberg}]{Ravfogel2021-jk}
Shauli Ravfogel, Grusha Prasad, Tal Linzen, and Yoav Goldberg. 2021.
\newblock \href {http://arxiv.org/abs/2105.06965} {Counterfactual interventions
  reveal the causal effect of relative clause representations on agreement
  prediction}.

\bibitem[{Schrimpf et~al.(2021)Schrimpf, Blank, Tuckute, Kauf, Hosseini,
  Kanwisher, Tenenbaum, and Fedorenko}]{Schrimpf2021-gy}
Martin Schrimpf, Idan~Asher Blank, Greta Tuckute, Carina Kauf, Eghbal~A
  Hosseini, Nancy Kanwisher, Joshua~B Tenenbaum, and Evelina Fedorenko. 2021.
\newblock The neural architecture of language: Integrative modeling converges
  on predictive processing.
\newblock \emph{Proc. Natl. Acad. Sci. U. S. A.}, 118(45).

\bibitem[{Stern et~al.(2017)Stern, Fried, and Klein}]{Stern2017-vm}
Mitchell Stern, Daniel Fried, and Dan Klein. 2017.
\newblock Effective inference for generative neural parsing.
\newblock In \emph{Proceedings of the 2017 Conference on Empirical Methods in
  Natural Language Processing}, pages 1695--1700, Copenhagen, Denmark.
  Association for Computational Linguistics.

\bibitem[{Tenney et~al.(2019)Tenney, Xia, Chen, Wang, Poliak, Thomas~McCoy,
  Kim, Van~Durme, Bowman, Das, and Pavlick}]{Tenney2019-al}
Ian Tenney, Patrick Xia, Berlin Chen, Alex Wang, Adam Poliak, R~Thomas~McCoy,
  Najoung Kim, Benjamin Van~Durme, Samuel~R Bowman, Dipanjan Das, and Ellie
  Pavlick. 2019.
\newblock \href {http://arxiv.org/abs/1905.06316} {What do you learn from
  context? probing for sentence structure in contextualized word
  representations}.

\bibitem[{Tucker et~al.(2022)Tucker, Eisape, Qian, Levy, and
  Shah}]{Tucker2022-yd}
Mycal Tucker, Tiwalayo Eisape, Peng Qian, Roger Levy, and Julie Shah. 2022.
\newblock \href {http://arxiv.org/abs/2204.09722} {When does syntax mediate
  neural language model performance? evidence from dropout probes}.

\bibitem[{Tucker et~al.(2021)Tucker, Qian, and Levy}]{Tucker2021-sj}
Mycal Tucker, Peng Qian, and Roger Levy. 2021.
\newblock What if this modified that? syntactic interventions with
  counterfactual embeddings.
\newblock In \emph{Findings of the Association for Computational Linguistics:
  {ACL-IJCNLP} 2021}, Stroudsburg, PA, USA. Association for Computational
  Linguistics.

\bibitem[{Warstadt and Bowman(2020)}]{Warstadt2020-js}
Alex Warstadt and Samuel~R Bowman. 2020.
\newblock \href {http://arxiv.org/abs/2007.06761} {Can neural networks acquire
  a structural bias from raw linguistic data?}

\bibitem[{Wilcox et~al.(2019)Wilcox, Qian, Futrell, Ballesteros, and
  Levy}]{Wilcox2019-lo}
Ethan Wilcox, Peng Qian, Richard Futrell, Miguel Ballesteros, and Roger Levy.
  2019.
\newblock \href {http://arxiv.org/abs/1903.00943} {Structural supervision
  improves learning of {Non-Local} grammatical dependencies}.

\bibitem[{Wilcox et~al.(2021)Wilcox, Vani, and Levy}]{Gotlieb_Wilcox2021-qx}
Ethan Wilcox, Pranali Vani, and Roger~P Levy. 2021.
\newblock A targeted assessment of incremental processing in neural
  {LanguageModels} and humans.
\newblock \emph{arXiv e-prints}, page arXiv:2106.03232.

\bibitem[{Wilcox et~al.(2020)Wilcox, Gauthier, Hu, Qian, and
  Levy}]{Wilcox2020-uo}
Ethan~Gotlieb Wilcox, Jon Gauthier, Jennifer Hu, Peng Qian, and Roger Levy.
  2020.
\newblock \href {http://arxiv.org/abs/2006.01912} {On the predictive power of
  neural language models for human {Real-Time} comprehension behavior}.

\bibitem[{Wolf et~al.(2019)Wolf, Debut, Sanh, Chaumond, Delangue, Moi, Cistac,
  Rault, Louf, Funtowicz, Davison, Shleifer, von Platen, Ma, Jernite, Plu, Xu,
  Le~Scao, Gugger, Drame, Lhoest, and Rush}]{Wolf2019-vb}
Thomas Wolf, Lysandre Debut, Victor Sanh, Julien Chaumond, Clement Delangue,
  Anthony Moi, Pierric Cistac, Tim Rault, R{\'e}mi Louf, Morgan Funtowicz, Joe
  Davison, Sam Shleifer, Patrick von Platen, Clara Ma, Yacine Jernite, Julien
  Plu, Canwen Xu, Teven Le~Scao, Sylvain Gugger, Mariama Drame, Quentin Lhoest,
  and Alexander~M Rush. 2019.
\newblock \href {http://arxiv.org/abs/1910.03771} {{HuggingFace's}
  transformers: State-of-the-art natural language processing}.

\end{thebibliography}
\bibliographystyle{acl_natbib}

\appendix

\section{Appendix}
\subsection{Training}
\label{sec:app-training}
Our probes are built in PyTorch \cite{Paszke_undated-ly}. All probes were trained with Adam \cite{Kingma2014-nr} with a learning rate of $10^{-3}$ except for the GAP, which used a learning rate of $10^{-5}$ after pretraining on the regression task of \citet{Hewitt2019-yb}. For multilayer architectures, a dropout rate of 0.2 was used between layers (3 layers), and hidden state size was fixed to be the same size as the model emission. The attentive probe was implemented using a GRU of 200 hidden units, biaffine attention, and a 3 layer MLP read-out network. All architectures were optimized in PyTorch lightning \cite{falcon2019pytorch}. Each model trained for at most 40 each but used early stopping with a plateau tolerance of 3. 

\subsection{Counterfactual details}
\label{sec:app-cfx}
\begin{figure*}[t!]
	\centering
    \includegraphics[width=\linewidth, trim={0 0cm 0 0cm},clip]{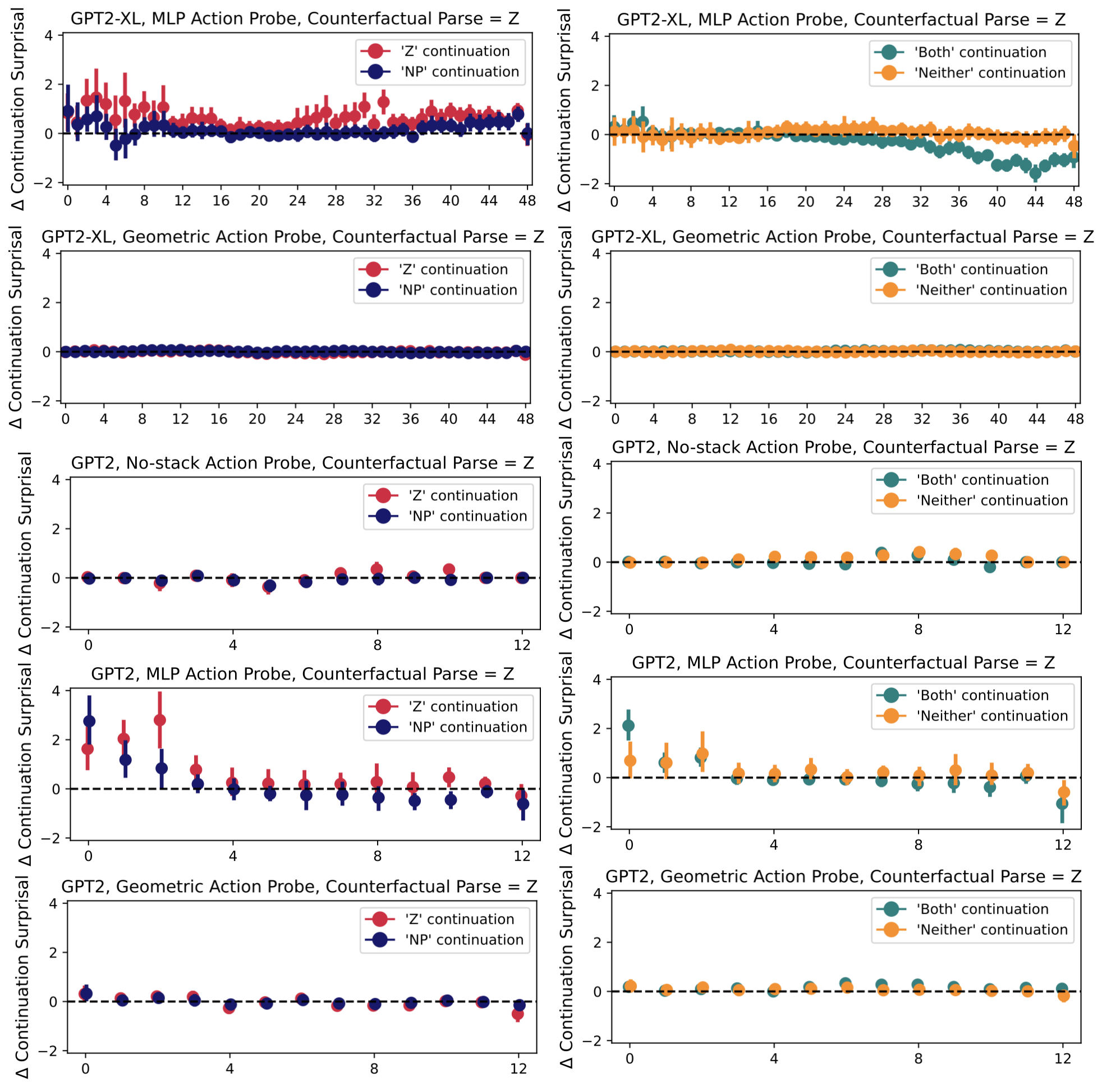}
    \caption{Extended results of counterfactual analyses.}
	\label{fig:supp_cfx1}
\end{figure*}

\begin{figure*}[t!]
	\centering
    \includegraphics[width=\linewidth, trim={0 0cm 0 0cm},clip]{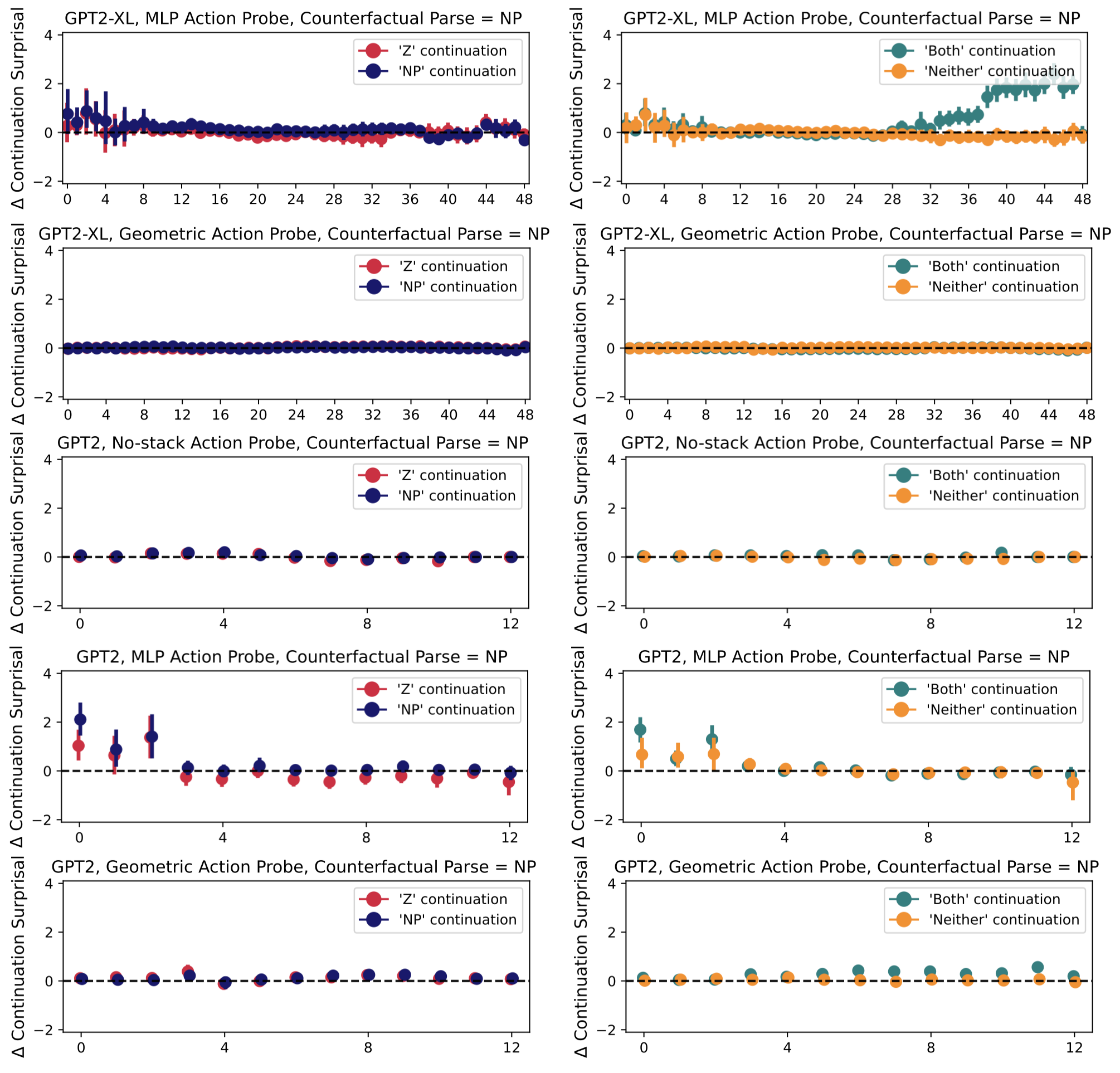}
    \caption{Extended results of counterfactual analyses}
	\label{fig:supp_cfx2}
\end{figure*}

We find training our probes with a high dropout rate \cite{Tucker2022-yd} improves counterfactual results. Thus, we train separate checkpoints for evaluation (0 dropout on model embeddings) and counterfactual generation (0.4 dropout). We found MAP achieved the largest counterfactual effect size in the predicted direction when evaluating NP- and Z-parse consistent completions. It had less interpretable effects on the `Both' and 'Neither' conditions. We include the full range of experiments in Figures \ref{fig:supp_cfx1} and \ref{fig:supp_cfx2}.


\end{document}